\documentclass[sigconf,natbib=true,anonymous=false]{acmart}
\AtBeginDocument{%
  \providecommand\BibTeX{{%
    \normalfont B\kern-0.5em{\scshape i\kern-0.25em b}\kern-0.8em\TeX}}}

\usepackage{multirow}

\setcopyright{acmcopyright}
\copyrightyear{2021}
\acmYear{2021}
\acmDOI{10.1145/1122445.1122456}

\acmConference[x]{x}{Nov 01--05, 2021}{Queensland, Australia}
\acmBooktitle{x}
\acmPrice{0}
\acmISBN{x}

\begin{document}

\title{MedGPT: Medical Concept Prediction from Clinical Narratives}
% Transformer based model captures medical knowledge from 600k pts

\author{Zeljko Kraljevic}
\email{zeljko.kraljevic@kcl.ac.uk}
\orcid{https://orcid.org/0000-0002-2310-2486}
\affiliation{%
  \institution{Kings College London}
  \streetaddress{Denmark Hill}
  \city{London}
  \country{United Kingdom}
}

\author{Anthony Shek}
\email{anthony.shek@kcl.ac.uk}
\orcid{https://orcid.org/0000-0002-7378-1261}
\affiliation{%
  \institution{Kings College London}
  \streetaddress{Denmark Hill}
  \city{London}
  \country{United Kingdom}
}

\author{Daniel Bean}
\email{daniel.bean@kcl.ac.uk}
\orcid{https://orcid.org/0000-0002-8594-7804}
\affiliation{%
  \institution{Kings College London}
  \streetaddress{Denmark Hill}
  \city{London}
  \country{United Kingdom}
}

\author{Rebecca Bendayan}
\email{rebecca.bendayan@kcl.ac.uk }
\orcid{https://orcid.org/0000-0003-1461-556X}
\affiliation{%
  \institution{Kings College London}
  \streetaddress{Denmark Hill}
  \city{London}
  \country{United Kingdom}
}

\author{James T.H. Teo}
\email{jamesteo@nhs.net}
\authornote{Both authors contributed equally}
\orcid{https://orcid.org/0000-0002-6899-8319}
\affiliation{%
  \institution{Kings College Hospital NHS Foundation Trust}
  \streetaddress{Denmark Hill}
  \city{London}
  \country{United Kingdom}
}

\author{Richard J.B. Dobson}
\email{richard.j.dobson@kcl.ac.uk}
\orcid{https://orcid.org/0000-0003-4224-9245}
\affiliation{%
  \institution{Kings College London}
  \streetaddress{Denmark Hill}
  \city{London}
  \country{United Kingdom}
}
\authornotemark[1]

\renewcommand{\shortauthors}{Kraljevic, et al.}

\begin{abstract}
The data available in Electronic Health Records (EHRs) provides the opportunity to transform care, and the best way to provide better care for one patient is through learning from the data available on all other patients. Temporal modelling of a patient's medical history, which takes into account the sequence of past events, can be used to predict future events such as a diagnosis of a new disorder or complication of a previous or existing disorder. While most prediction approaches use mostly the structured data in EHRs or a subset of single-domain predictions and outcomes, we present MedGPT a novel transformer-based pipeline that uses Named Entity Recognition and Linking tools (i.e. MedCAT) to structure and organize the free text portion of EHRs and anticipate a range of future medical events (initially disorders). Since a large portion of EHR data is in text form, such an approach benefits from a granular and detailed view of a patient while introducing modest additional noise. MedGPT effectively deals with the noise and the added granularity, and achieves a precision of 0.344, 0.552 and 0.640 (vs LSTM 0.329, 0.538 and 0.633) when predicting the  top 1, 3 and 5 candidate future disorders on real world hospital data from King's College Hospital, London, UK (\textasciitilde600k patients). We also show that our model captures medical knowledge by testing it on an experimental medical multiple choice question answering task, and by examining the attentional focus of the model using gradient-based saliency methods.
\end{abstract}




\begin{teaserfigure}
  \includegraphics[width=\textwidth]{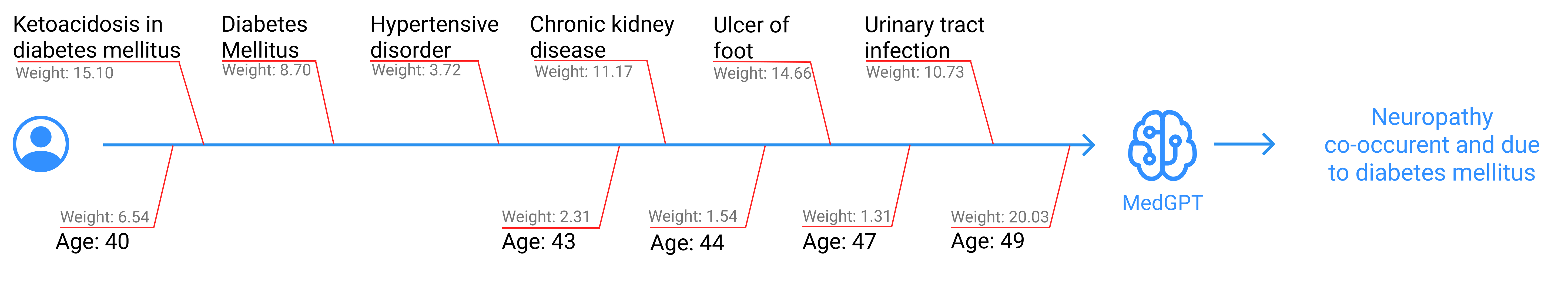}
  \caption{Importance of each token in the patient timeline for prediction of the right-most disorder using MedGPT. The weight was calculated using gradient-based saliency methods.}
  \Description{Importance of each token in the patient timeline for prediction of the right-most disorder using MedGPT. The weight was calculated using gradient-based saliency methods.}
  \label{fig:teaser}
\end{teaserfigure}

\maketitle

\section{Introduction and Related Work}

Electronic Health Records (EHRs) hold detailed longitudinal information about each patients' health status and disease progression, the majority of which are stored within unstructured text. Temporal models utilising such data could be used to predict future events such as a diagnosis, illness trajectory, risk of procedural complications, or medication side-effects. The majority of previous work for prediction or forecasting uses structured datasets or structured data in EHRs. Recently, numerous attempts have been made using BERT-based models. Examples include BEHRT \cite{BEHRT} which uses a limited subset of disorders (301 in total) available in the structured portion of EHRs. BEHRT is limited to predictions of disorders occurring in the next patient hospital visit or a specific predefined time frame, consequently, requiring that the information is grouped into patient visits. In addition, we note that the approach is a multi-label approach, which can cause difficulties as the number of concepts to be predicted increases. Another example is G-BERT \cite{GBERT}, the inputs for this model are all single-visit samples, which are insufficient to capture long-term contextual information in the EHR. Similarly to BEHRT, only the structured data is used. Next, Med-BERT \cite{MedBERT} is trained on structured diagnosis data, coded using the International Classification of Diseases. The model is not directly trained on the target task of predicting a new disorder, but fine-tuned after the standard Masked Language Modeling (MLM) task. The model is evaluated on a small subset of disorders which may be insufficient for estimating general performance.
Apart from BERT based models, we also note Long Short Term Memory (LSTM) models, like the one proposed by Ethan Steinberg et al. \cite{lmLSTM}. Similar to the other models, they only use structured data and fine-tune their model for the prediction of limited future events.

Most deep learning models concerned with predicting a wide range of future medical events tend to focus on the structured portion of EHRs. A large proportion of real world data however is unstructured, uncurated and often has information contained in richer clinical narratives \cite{cogstack}. Transforming unstructured data into a machine process-able sequential structured form provides a framework for forecasting the next medical concept in sequence, with a significant increase in granularity and applicability.

In this work we reuse the free text data within the EHR and build a general purpose model that is directly trained on the target task. This work, to some extent, follows the approach outlined in GPTv3 \cite{gpt3} and is similar to architectures where different tasks are implicit in the dataset, as an example, one GPTv3 model can generate HTML code, answer questions, write stories and much more without any fine-tuning.

\section{Methods}

MedGPT is a transformer-based pipeline for medical concept forecasting from clinical narratives. It is built on top of the GPTv2 \cite{gpt2} architecture which allows us to do causal language modeling (CLM). EHR data is sequentially ordered in time and this sequential order is important \cite{time}. As such Masked Language Modeling (MLM) approaches like BERT \cite{bert}, are not a good fit because when predicting the masked token, BERT models can also look into the future. Formally the task at hand can be defined as: given a corpus of patients $U = \{u_1, u_2, u_3, ..\}$ where each patient is defined as a sequence of tokens $u_i = \{w_1, w_2, w_3, ...\}$ and each token is medically relevant and temporally defined piece of patient data, our objective is the standard language modeling objective:
$$
L(U) = \sum_i \sum_j log P(w_j^i|w_{j-1}^i, w_{j-2}^i, ... w_0^i) 
$$
Note that in this work each of the tokens $w_i$ represents either the patient's age or a SNOMED-CT disorder concept that relates to the patient and is not negated.

\subsection{Named Entity Recognition and Linking}

The Medical Concept Annotation Toolkit (MedCAT \cite{medcat}) was used to extract disorder concepts from free text and link them to the SNOMED-CT concept database. MedCAT is a set of decoupled technologies for developing Information Extraction (IE) pipelines for varied health informatics use cases. It uses self-supervised learning to train a Named Entity Recognition and Linking (NER+L) model for any concept database (in our case SNOMED-CT) and demonstrates state-of-the-art performance. In addition to NER+L, MedCAT also supports concept contextualization with supervised training e.g. Negation detection (is the concept negated or not). 

We applied a concept filter to MedCAT and used it to extract only concepts that are marked as \textit{disorders} in SNOMED-CT (in total 77265 disorders). For the meta-annotations we trained (supervised) two models: Negation and Subject (is the extracted concept related to the patient or someone/something else).

MedCATtrainer \cite{medcattrainer} was used to train the two supervised models for contextualization and to provide supervised `top up' training for the unsupervised NER+L model. We picked the top 100 most frequent disorders in the dataset and sampled 2 documents for each in which they occur. We also sampled 100 random documents from the whole dataset to avoid biasing the training to the most frequent disorders. These 300 documents were first annotated by MedCAT and then manually validated for missing and incorrect annotations, in total 12668 annotations. Annotations were done by AS and ZK predominantly, with clinical annotators helping out. Annotators followed annotation guidelines to keep consistency with additional periodic annotation counter-checking to resolve uncertainties.

\subsection{Exploratory Analysis of Modifications for Transformer Based Models}
Transformer-based models are currently one of the most popular architectures for deep learning, as a result there is a significant amount of proposed modifications to improve performance. We performed an exploratory analysis of 8 different approaches on top of the base GPTv2 model.

The modifications tested were: 1) Memory Transformers \cite{memory-transformer} - Mikhail S. Burtsev et al. introduced memory tokens and showed that adding trainable memory to selectively store local as well as global representations of a sequence is a promising direction to improve the Transformer model. We used the base version of the memory transformer with 20 memory tokens. 2) Residual Attention \cite{residual-attention} -  Ruining He et al. showed a simple Residual Attention Layer Transformer architecture that significantly outperformed canonical Transformers on a spectrum of tasks. 3) ReZero \cite{rezero} - Thomas Bachlechner et al. showed that a simple architecture change of gating residual connections using a single zero-initialized parameter improved the convergence speed for deep networks. 4) Talking Heads Attention \cite{talking-heads} - Noam Shazeer et al. introduced a variation on multi-head attention which included linear projections across the attention-heads dimension, immediately before and after the softmax operation. 5) Sparse Transformers \cite{sparse-transformers} - Guangxiang Zhao et al. demonstrated that it is possible to improve the concentration of attention on the global context through an explicit selection of the most relevant segments. 6) Rotary embeddings \cite{rotary} - Jianlin Su et al. proposed to encode absolute positional information with a rotation matrix and incorporate explicit relative position dependency in self-attention formulation. The approach showed promising results on long sequences. 7) GLU \cite{glu} - Noam Shazeer showed that Gated Linear Unites yield quality improvements over standard ReLU or GELU activation functions in transformer based models. 8) Word2Vec - We tested one additional approach where we initialized the transformer token embeddings with pre-calculated embeddings from MedCAT. 

Implementations of the modifications were either taken from HuggingFace Transformers \cite{huggingface}, x-transformers\footnote{https://github.com/lucidrains/x-transformers/} or from the repository published by the authors where applicable.

\subsection{Dataset Preparation}

Two EHR datasets were used: King's College Hospital (KCH) NHS Foundation Trust, UK and MIMIC-III \cite{mimiciii}. No preprocessing or filtering was done on the MIMIC-III dataset of clinical notes and all 2083179 free text documents were used directly. At KCH we collected a total of 18436789 documents (all clinical activity on EHR from 1999 till January 2021), retrieved from the EHR using CogStack \cite{cogstack}. We removed all documents that were of bad quality (OCR-related problems) or where there may be ambiguity in presence of disorder (e.g. incomplete triage checklists, questionnaires and forms). After this filtering step we were left with 13084498 documents, each of which had a timestamp representing the time when the document was written. Some documents are continuous - meaning more information is added to them over time, these were split into fragments where each was defined with a time-of-writing. The project operated under London South East Research Ethics Committee (reference 18/LO/2048) approval granted to the King’s Electronic Records Research Interface (KERRI); specific approval in using NLP on unstructured clinical text for extraction of standardised biomedical Codes for patient record structuring was reviewed with expert patient input on a virtual committee with Caldicott Guardian oversight.

Using MedCAT we extracted SNOMED-CT concepts representing disorders attributed to the patient and those that are not negated (based on MedCAT meta-annotations) from both datasets as described above. The concepts were then grouped by patient and only the first occurrence of a concept was kept. Another filtering step was applied to increase the confidence of a diagnosis, namely a concept was kept if it appeared at least twice in the patients EHR. This was then used to create a sequence of pathology for each patient, as shown in Figure \ref{fig:teaser}. Each concept was prepended with patient age, only if the age had changed since the last disorder in the sequence for that patient. 

Without any filtering we had 1121218 patients at KCH and 42339 at MIMIC-III, after removal of all disorders with frequency < 100 and all patients that have < 5 tokens we were left with 582548 and 33975 patients respectively. For this work we also limited the length of each sample/patient to 50 tokens (Both KCH and MIMIC-III have more than 98\% of patients with less than 50 tokens). The resulting dataset was then split into a train/test set with an 80/20 ratio. The train set was further split into a train/validation set with a 90/10 ratio. The validation set was used for hyperparameter tuning. All presented scores are calculated on the test set. All sampling was random. 

\section{Results}

The performance of MedCAT on our annotated dataset consisting of 12668 annotations was $F1=0.93$ for NER+L, for the meta-annotation Subject $F1=0.95$ and for the meta-annotation Negation $F1=0.91$. All metrics were calculated on the test set (10\% of the total 12668 annotations). 

The MedGPT transformer model is built on-top of the GPTv2 architecture. To find the optimal parameters for the base GPTv2 model on our dataset we used Population Based Training \cite{population-training}, the best result was achieved with \textit{n\_layers=6}, \textit{n\_attention\_heads=2}, \textit{embedding\_dim=300}, \textit{weight\_decay=0.14}, \textit{lr=4.46e-5}, \newline \textit{batch\_size=32} and \textit{warmup\_steps=15}. 

\subsection{Exploratory Analysis of Modifications for Transformer Based Models}

To improve the base GPTv2 model, we undertook an exploratory analysis on 8 recent improvements for transformer based models. As the baseline we used the best model obtained after hyperparameter tuning of the base GPTv2 model, note that we did not do any additional hyperparameter tuning in this step, but only used the parameters obtained for the base model. The two most promising approaches were in the end combined to achieve the best results (see Table \ref{tab:modifications}).

\begin{table}[]
\begin{tabular}{llllll} \toprule
\textbf{Model} \textbackslash \textbf{Dataset} & \multicolumn{5}{c}{\textbf{KCH}} \\
\hline
                                     & P @1    & P@3     & P @5    & H 10+   & H 20+  \\
Base GPT                             & 0.342   & 0.550   & 0.639   & 0.380   & 0.386   \\
Memory Tokens 20                     & 0.341   & 0.547   & 0.636   & 0.376   & 0.381  \\
Residual Attention                   & 0.337   & 0.543   & 0.632   & 0.370   & 0.375  \\
ReZero                               & 0.307   & 0.50    & 0.603   & 0.327   & 0.333  \\
Talking Heads                        & 0.342   & 0.550   & 0.638   & 0.381   & 0.387  \\
Sparse Top 8                         & 0.341   & 0.548   & 0.638   & 0.380   & 0.384  \\
Rotary                               & 0.342   & 0.548   & 0.639   & 0.382   & \textbf{0.389}  \\
GLU                                 & 0.343   & 0.550   & \textbf{0.640}   & \textbf{0.383}   & \textbf{0.389}  \\
Word2Vec                             & 0.342   & 0.550   & \textbf{0.640}   & 0.380   & 0.386  \\
GLU + Rotary                        & \textbf{0.344}   & \textbf{0.551}   & \textbf{0.640}   & \textbf{0.383}   & \textbf{0.389} \\
\bottomrule 
\end{tabular}
\caption{Precision for next disorder prediction calculated on the dataset from King's College Hospital. Here $@N$ means that out of $N$ candidates predicted by the model at least one is correct. And $H \, k+$ is Precision calculated only for disorders appearing at position $k+$.}
\label{tab:modifications}
\end{table}

\subsection{MedGPT Next Disorder Prediction}
The MedGPT model, which consists of the GPTv2 base model with the GLU+Rotary extension, is tested on two datasets KCH and MIMIC-III for the task of predicting the next disorder in a patient's timeline. We compared our model to a standard Bag of Concepts (BoC) approach with a SVM classifier, as well as a Long Short Term Memory (LSTM) network (Table \ref{tab:precision} and \ref{tab:hprecision}).  

\begin{table}[]
\begin{tabular}{llll|lll} \toprule
 & \multicolumn{3}{c}{\textbf{MIMIC-III}} & \multicolumn{3}{c}{\textbf{KCH}} \\
\hline
                 & P @1    & P@3     & P @5   & P @1    & P@3     & P @5 \\
\hline
BoC SVM          & 0.331   & -       & -      & 0.215   & -       & -   \\
LSTM             & 0.419   & 0.657   & 0.746  & 0.329   & 0.538   & 0.633  \\
MedGPT           & \textbf{0.443}    & \textbf{0.681}   & \textbf{0.770}     & \textbf{0.344}   & \textbf{0.551}    & \textbf{0.640}  \\
\bottomrule 
\end{tabular}
\caption{Precision for next disorder prediction calculated on patients from the MIMIC-III and King's College Hospital. Here $@N$ means that out of the $N$ candidates predicted by the model at least one of them is correct.}
\label{tab:precision}
\end{table}
\iffalse
\begin{table*}[]
\begin{tabular}{lllll|llll} \toprule
\textbf{Model} \textbackslash \textbf{Dataset} & \multicolumn{4}{c}{\textbf{MIMIC-III}} & \multicolumn{4}{c}{\textbf{KCH}} \\
\hline
& \multicolumn{8}{c}{ \textbf{Precision} } \\
                 & H 0+  & H 10+   & H 20+  & H 30+.     & H 0+  & H 10+   & H 20+  & H 30+.  \\
                 \hline
LSTM             & 0.419     & 0.394   & 0.367  & 0.338      & 0.329   & 0.367   & 0.371  & 0.365  \\
BoC SVM          & 0.331     & 0.335   & 0.319  & 0.293      & 0.215   & 0.250   & 0.248   & 0.233  \\
MedGPT           & \textbf{0.443}  & \textbf{0.431}   & \textbf{0.411}  & \textbf{0.392}      & \textbf{0.344}   & \textbf{0.383}   & \textbf{0.389}   & \textbf{0.386}  \\
\hline
Support (test set) & 6795 & 4244 & 2048 & 1068        & 116510   & 58323 & 22867 & 11925 \\
\bottomrule 
\end{tabular}
\caption{Precision calculated only for disorders appearing at position 0+, 10+, 20+ or 30+ in a patient's timeline. This shows the performance of the model with respect to different amounts of historical information.}
\label{tab:hprecision}
\end{table*}
\fi

\begin{table}[]
\begin{tabular}{lllll} \toprule
\textbf{Model} \textbackslash \textbf{Dataset} & \multicolumn{4}{c}{\textbf{MIMIC-III}} \\
\hline
                 & H 0+  & H 10+   & H 20+  & H 30+.     \\
                 \hline
LSTM             & 0.419     & 0.394   & 0.367  & 0.338    \\
BoC SVM          & 0.331     & 0.335   & 0.319  & 0.293  \\
MedGPT           & \textbf{0.443}  & \textbf{0.431}   & \textbf{0.411}  & \textbf{0.392} \\
\hline
Support (test set) & 6795 & 4244 & 2048 & 1068      \\
\bottomrule 
 & \multicolumn{4}{c}{\textbf{KCH}} \\
\hline
LSTM & 0.329   & 0.367   & 0.371  & 0.365  \\
BoC SVM & 0.215   & 0.250   & 0.248   & 0.233 \\
MedGPT & \textbf{0.344}   & \textbf{0.383}   & \textbf{0.389}   & \textbf{0.386} \\
\hline
Support (test set) &  116510   & 58323 & 22867 & 11925 \\
\bottomrule
\end{tabular}
\caption{Precision calculated only for disorders appearing at position 0+, 10+, 20+ or 30+ in a patient's timeline. This shows the performance of the models with respect to different amounts of historical information.}
\label{tab:hprecision}
\end{table}

\subsection{Qualitative Analysis and Interpretability}
To explore the model's capabilities a senior clinician crafted 4 Multiple Choice Questions (MCQ) set to challenge the model. We present the model with an imaginary patient until a time point $t$ and ask a MCQ (in all cases there is a medical explanation why one answer should be more likely than the others). As MedGPT calculates the probability of all tokens in the vocabulary when predicting the next disorder, we can display the probability of the disorders in question. Note that the probabilities for the options were normalized and that it is possible that they are not the top predictions by the model. Figure \ref{tab:mcq} shows the 4 examples and the decision made by MedGPT, and even though the model was not directly trained on a ranking task its predictions align with expert clinical knowledge. 

\textbf{Example 1:} As a first test, a high-level term of Diabetes Mellitus was provided in the context of a subtype-specific complication, and we tried to see if it could determine which category of diabetes mellitus is most associated with ketoacidosis. This is a simple binary task which it performed well, consistent with medical literature. 

\textbf{Example 2:} For this, a choice of a common condition (diabetic nephropathy) was provided as a distracting choice to a more uncommon scenario (congenital cystic kidney disease). In this scenario, the background (cerebral aneurysm) provided the contextual cue for the rarer diagnosis which MedGPT successfully discerned. 

\textbf{Example 3:} To test the longer-attention, this scenario introduced the main pertaining disorders early in the sequence (Psychotic disorder, Bipolar disorder, Schizoaffective disorder). Several other disorders (seizure, epilepsy, ischemic heart disease, hypertensive disorder) were introduced late in the sequence to see if this would disrupt the most likely prediction. MedGPT also successfully handled the necessary indirect inference that the drug treatments which cause the diagnosis were not explicitly stated either.

\textbf{Example 4:} Similar to above Example 3, attention in the presence of distractors were tested through intermixing historical diseases. Primary Sclerosing Cholangitis is the most relevant premorbid diagnosis as it is associated with inflammatory bowel diseases (Crohn's Disease and Ulcerative Colitis) and the MedGPT was able to distinguish this from several other common conditions.

\begin{figure}[]
\begin{tabular}{|c|p{0.4\linewidth}|p{0.33\linewidth}|p{0.09\linewidth}|} \hline
& \textbf{Patient history} & \textbf{Multiple Choice Options} & \textbf{P} \\
\hline
1 & 40 -> Ketoacidosis in Diabetes Mellitus -> Diabetes Mellitus -> Hypertension  & T1 Diabetes Mellitus \newline T2 Diabetes Mellitus & \textbf{0.92} \newline 0.08 \\ 
\hline
2 & 38 -> Hypertensive disorder -> 41 -> Chronic kidney disease -> 43 -> Subarachnoid haemorrhage -> 44 -> Cerebral aneurysm -> 46 -> Microscopic haematuria -> 48 & Congenital cystic kidney disease \newline Renal Artery Stenosis \newline Diabetic Nephropathy & 
                        \textbf{0.8} \newline \newline 0.175 \newline 0.025 \\
 \hline
3 & 21 -> Psychotic disorder -> 24 -> Bipolar disorder -> 28 -> Schizoaffective disorder -> 35 Depressive disorder -> 42 -> Hypertensive disorder -> 44 -> Seizure disorder -> 49 Epilepsy -> 55 -> Ischemic heart disease -> 58 Mild cognitive disorder -> 59 & 
Parkinsonism caused by drug \newline Drug-induced tardive dyskinesia \newline Vascular dementia \newline Alzheimer's disease \newline Parkinson's disease 
& \textbf{0.56} \newline \newline 0.39 \newline \newline 0.019 \newline 0.018 \newline 0.013 \\
\hline
4 & 41 -> Gastroesophageal reflux disease -> 46 -> Cholestatic jaundice syndrome -> Pancreatitis -> 47 -> Primary sclerosing cholangitis -> 51 -> Acute diarrhea -> 53 -> Pancreatitis -> 55 -> Hemorrhagic diarrhea -> 56 & 
Crohn's disease \newline Ulcerative Colitis \newline Diverticulitis \newline Haemorrhoids \newline Rectal Adenocarcinoma 
& \textbf{0.52} \newline 0.29 \newline 0.12 \newline 0.04 \newline 0.03 \\
\hline 
\end{tabular}
\caption{Qualitative analysis of MedGPT on 4 multiple choice questions. In all cases the model was given the patient history and asked to predict the probability of each of the options. The P column denotes the normalized probability assigned by MedGPT.}
\label{tab:mcq}
\end{figure}

\smallskip
To understand why a certain disorder was forecasted we used gradient-based saliency methods \cite{gradsal}. This method allowed us to calculate how important each input token was for the prediction of the next disorder in sequence. We show the potential of the model in Figure \ref{fig:teaser}. Age (49) was the most salient input token followed by the tokens \textit{ketoacidosis} and \textit{ulcer of foot}. These disorder tokens are clearly very relevant as the \textit{ketoacidosis} concept contextualises to patients with \textit{diabetes mellitus} and the \textit{ulcer} suggests sub-clinical nerve disease which is common in patients with diabetes. Age is likely to be relevant in most forecasting scenarios. Such weights allow probing the forecast with both qualitative and quantitative interpretations. The potential to provide explainability makes this distinct from current commercial apps using structured question trees that have modest real-world performance metrics \cite{mcq-apps}.

\section{Conclusion}
We demonstrate that unstructured clinical narratives can be used for temporal modelling and forecasting of medical history through a multi-stage process where unstructured data is first parameterised using NER+L into a standardised ontology, then a second step of Transformer-based NLP is applied for forecasting. MedGPT shows potential to overcome real-world scenarios where electronic health records exist at various levels of maturity in data standardisation.

Additionally, MedGPT's promising ability to choose from a set of differential diagnoses without further training suggests that it has captured relational associations between medical disorders and is able to focus attention to salient parts of the chronology to do so. This will be tested more extensively in future work.

\clearpage
\bibliographystyle{ACM-Reference-Format}
\bibliography{medflux}

\appendix

\end{document}